\documentclass[runningheads]{llncs}
\usepackage[T1]{fontenc}
\usepackage[protrusion=true,expansion=false]{microtype}
\usepackage{graphicx}
\usepackage{booktabs}
\usepackage{multirow}
\usepackage{amsmath,amssymb}
\usepackage[table,dvipsnames]{xcolor}

\begin{document}

\title{INST-Align: Implicit Neural Alignment for Spatial Transcriptomics via Canonical Expression Fields}
\titlerunning{INST-Align: Implicit Neural Alignment for Spatial Transcriptomics}

\author{Bonian Han \and Cong Qi \and Przemyslaw Musialski \and Zhi Wei}
\authorrunning{Han et al.}
\institute{New Jersey Institute of Technology, Newark, NJ, USA\\
\email{\{bh348, cq5, przemyslaw.musialski, zhi.wei\}@njit.edu}}
\maketitle

\begin{abstract}
Spatial transcriptomics (ST) measures mRNA expression while preserving spatial organization, but multi-slice analysis faces two coupled difficulties: large non-rigid deformations across slices and inter-slice batch effects when alignment and integration are treated independently. We present INST-Align, an unsupervised pairwise framework that couples a coordinate-based deformation network with a shared Canonical Expression Field, an implicit neural representation mapping spatial coordinates to expression embeddings, for joint alignment and reconstruction. A two-phase training strategy first establishes a stable canonical embedding space and then jointly optimizes deformation and spatial-feature matching, enabling mutually constrained alignment and representation learning. Cross-slice parameter sharing of the canonical field regularizes ambiguous correspondences and absorbs batch variation. Across nine datasets, INST-Align achieves state-of-the-art mean OT Accuracy (0.702), NN Accuracy (0.719), and Chamfer distance, with Chamfer reductions of up to 94.9\% on large-deformation sections relative to the strongest baseline. The framework also yields biologically meaningful spatial embeddings and coherent 3D tissue reconstruction. The code will be released after review phase.
\end{abstract}

\section{Introduction}

Spatial transcriptomics (ST) enables gene expression measurement with spatial context~\cite{ref_st,ref_stmrna}, spanning imaging-based approaches such as multiplexed FISH and in situ sequencing~\cite{ref_stmrna,ref_starmap} and sequencing-based spatial barcoding platforms~\cite{ref_st,ref_slideseqv2}. These technologies have enabled molecular mapping across diverse biological systems~\cite{ref_zhang_limb,ref_qi2025,ref_wang2025,ref_allen_brain,ref_khaliq_cancer,ref_longo_review}, from the human dorsolateral prefrontal cortex~\cite{ref_slideseq} to whole mouse embryos~\cite{ref_embryo}, making multi-slice tissue analysis increasingly practical. However, reconstructing coherent 3D tissue volumes from serial sections remains challenging~\cite{ref_benchmark}: slices undergo complex non-rigid distortions during sample preparation (tearing, folding, compression), and arbitrary inter-slice rotations further compound the geometric ambiguity. Biological complexity makes ground-truth correspondences largely unattainable, rendering alignment inherently unsupervised. Without a common coordinate frame, cross-slice expression integration is confined to individual 2D sections, leaving the collective three-dimensional context unexploited.

Existing alignment methods fall into two categories~\cite{ref_benchmark}. \emph{Integration-based} methods such as STAligner~\cite{ref_staligner}, GraphST~\cite{ref_graphst}, SPACEL~\cite{ref_spacel}, and SANTO~\cite{ref_santo} learn joint spatial embeddings to correct coordinates, but their alignment quality is bounded by representation fidelity. \emph{Geometric} methods align coordinates directly via optimal transport~\cite{ref_paste}, Gaussian processes~\cite{ref_qi2025,ref_gpsa}, or diffeomorphic mappings~\cite{ref_stalign,ref_spateo}, but treat alignment and integration as independent steps. A common limitation is that these approaches decouple spatial alignment from expression integration: integration-based methods risk over-correcting and diminishing genuine biological diversity~\cite{ref_benchmark}, while geometric methods lack expression-level feedback. Since raw expression inherently contains inter-slice batch effects, we argue that alignment and integration should be solved jointly.

Implicit Neural Representations (INRs) model signals as continuous coordinate-conditioned functions $f_\theta: \mathbf{x} \mapsto \mathbf{y}$, enabling resolution-independent querying with compact parameterizations~\cite{ref_nerf,ref_siren}. Although INRs are naturally suited to this setting---jointly learning deformation and reconstruction from coordinates, as demonstrated by Nerfies~\cite{ref_nerfies} in novel-view synthesis---and implicit deformation models have been explored in medical image registration~\cite{ref_cidir,ref_dio_reg}, existing INR-based ST methods such as STINR~\cite{ref_stinr} and SUICA~\cite{ref_suica} only model within-slice expression fields and do not estimate deformation fields for multi-slice alignment.

In this paper, we propose INST-Align, a framework that couples spatial alignment and expression integration through a shared Canonical Expression Field parameterized as an INR. Requiring all slices to be explained by a \emph{single continuous field} regularizes the ill-posed registration problem, turning alignment into a mutually constrained reconstruction task. Our key contributions are:
\begin{itemize}
\item \textbf{Joint alignment and integration via a shared canonical field.} A shared Canonical Expression Field coupled with a deformation network jointly solves spatial alignment and expression integration, achieving state-of-the-art mean alignment performance across nine datasets.
\item \textbf{Spatial-expression alignment loss.} A unified cost combining geometric proximity and canonical-field feature similarity with adaptive soft assignment drives unsupervised non-rigid registration without ground-truth correspondences.
\item \textbf{Biologically meaningful embeddings.} The canonical field produces spatial embeddings that support downstream analysis (e.g., clustering) without requiring dedicated graph-based embedding methods.
\end{itemize}

\section{Method}

\begin{figure}[t]
\includegraphics[width=\textwidth]{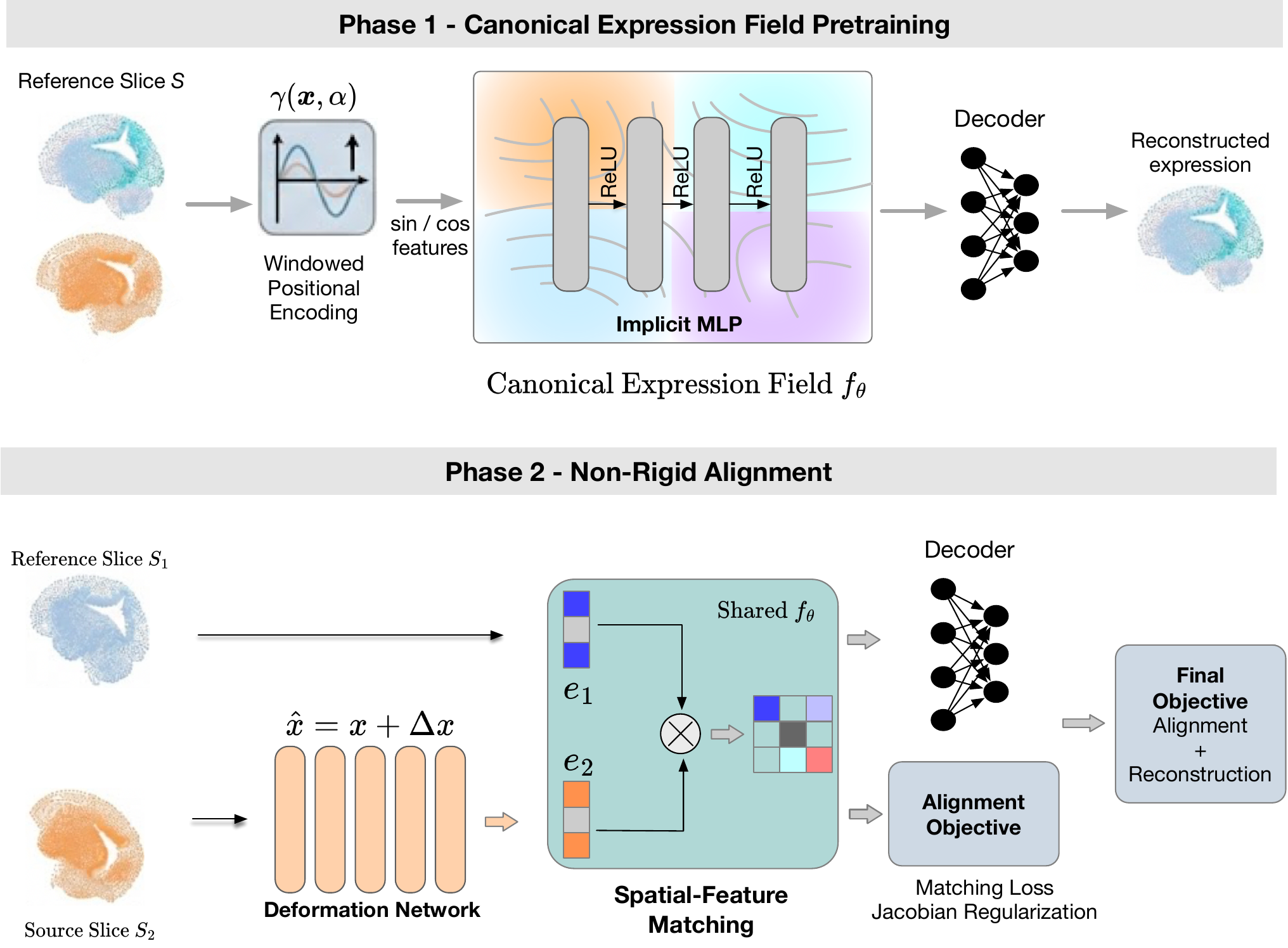}
\caption{Overview of the INST-Align pipeline. Phase 1 learns a shared canonical expression field $f_\theta$ as an implicit neural representation that maps spatial coordinates to embedding vectors and reconstructs gene expression profiles. Phase 2 jointly optimizes a non-rigid deformation network and feature-guided spatial matching within the shared field, enabling mutually reinforcing alignment and representation learning under smoothness regularization.} \label{fig:pipeline}
\end{figure}

Fig.~\ref{fig:pipeline} illustrates the overall pipeline of INST-Align. Given a pair of ST slices, we first perform rigid alignment via adaptive ICP, then jointly optimize a Canonical Expression Field and a Deformation Network through two-phase training. In Phase 1, we pretrain slice-specific expression fields with a shared decoder to establish a stable and biologically meaningful embedding space. In Phase 2, we switch to a single shared canonical field and jointly optimize deformation and alignment losses, enabling mutually constrained non-rigid registration and representation learning.

\subsection{Problem Formulation}

Let $\mathcal{S}_1 = \{(\mathbf{x}_i^{(1)}, \mathbf{g}_i^{(1)})\}_{i=1}^{N_1}$ and $\mathcal{S}_2 = \{(\mathbf{x}_j^{(2)}, \mathbf{g}_j^{(2)})\}_{j=1}^{N_2}$ denote two ST slices, where $\mathbf{x} \in \mathbb{R}^2$ are spatial coordinates and $\mathbf{g} \in \mathbb{R}^G$ are gene expression vectors. We designate $\mathcal{S}_1$ as the reference slice and $\mathcal{S}_2$ as the source slice.

Our goal is to learn: (1) a deformation function $\phi_\psi: \mathbb{R}^2 \rightarrow \mathbb{R}^2$ that maps source coordinates into the reference frame, and (2) a canonical expression field $f_{\theta_1}: \mathbb{R}^2 \rightarrow \mathbb{R}^d$ that maps any spatial location to a $d$-dimensional embedding capturing the biological identity at that position. The coordinate pairs are z-score normalized to a common scale, and a rigid pre-alignment is applied using adaptive ICP~\cite{ref_icp,ref_icp_survey} with expression-guided rotation selection.

\subsection{Canonical Expression Field}

The Canonical Expression Field is parameterized as an implicit neural representation (ExprINR) $f_{\theta_1}$ defined on the reference frame. Given a 2D coordinate $\mathbf{x}$, we first apply a windowed positional encoding~\cite{ref_nerfies,ref_fourier}:
\begin{equation}
\gamma(\mathbf{x}, \alpha) = \Big[\mathbf{x},\; w_k(\alpha) \sin(2\pi \sigma_k \mathbf{x}),\; w_k(\alpha) \cos(2\pi \sigma_k \mathbf{x}) \Big]_{k=0}^{L-1}
\label{eq:pe}
\end{equation}
where $\sigma_k = 2^{k \cdot l_{\max} / (L-1)}$ are log-linearly spaced frequencies and $w_k(\alpha)$ is the coarse-to-fine window of~\cite{ref_nerfies}. The window parameter $\alpha \in [0, L]$ increases linearly during training, progressively enabling higher frequencies to prevent overfitting to noise in early epochs.

The encoded coordinates are processed by a 4-layer MLP with LayerNorm and ReLU activations to produce a $d$-dimensional bottleneck embedding ($d=64$):
\begin{equation}
\mathbf{e} = f_\theta(\mathbf{x}, \alpha) = \text{MLP}_\theta(\gamma(\mathbf{x}, \alpha)) \in \mathbb{R}^d
\end{equation}

An expression decoder $h_\phi: \mathbb{R}^d \rightarrow \mathbb{R}^{G'}$ reconstructs the highly variable gene (HVG) expression from the embedding:
\begin{equation}
\hat{\mathbf{g}} = h_\phi(\mathbf{e})
\end{equation}

A key design choice spans both training phases. In Phase~1, two independent ExprINRs ($f_{\theta_1}$ for the reference, $f_{\theta_2}$ for the source) share a single decoder, forcing their embedding spaces to align. In Phase~2, the source INR ($f_{\theta_2}$) is discarded: since the Deformation Network maps source coordinates into the reference frame, a single canonical field $f_{\theta_1}$ can encode both slices. This progressive design---from paired to canonical---ensures the canonical field is well-initialized before joint optimization begins, providing a stable biological prior that guides deformation.

\subsection{Deformation Network}

The Deformation Network $\phi_\psi$ maps source coordinates to a displacement field. Given rigidly aligned source coordinates $\mathbf{x}^{(2)}$, it predicts:
\begin{equation}
\hat{\mathbf{x}}^{(2)} = \phi_\psi(\mathbf{x}^{(2)}, \alpha) = \mathbf{x}^{(2)} + \Delta\mathbf{x}(\mathbf{x}^{(2)})
\end{equation}
where $\Delta\mathbf{x}$ is the learned displacement. The network consists of a 6-layer trunk MLP ($d_{\text{hidden}}=128$) followed by a lightweight spatial head ($d_{\text{hidden}}=64$, 1 layer) that outputs the 2D displacement. The spatial head is initialized near zero so that $\phi_\psi$ starts as an identity mapping, ensuring stable early-stage optimization.

To ensure spatially smooth and physically plausible deformations, we regularize the Jacobian of $\phi_\psi$. For a sampled subset of points, we compute the $2 \times 2$ Jacobian matrix $\mathbf{J} = \partial \hat{\mathbf{x}} / \partial \mathbf{x}$ via automatic differentiation and extract its singular values $\sigma_1, \sigma_2$ via SVD. The regularization loss penalizes deviations from volume-preserving behavior with an \textbf{asymmetric} weighting:
\begin{equation}
\mathcal{L}_{\text{jac}} = \frac{1}{|\mathcal{B}|} \sum_{\mathbf{x} \in \mathcal{B}} \sum_{k=1}^{2} w_k \cdot (\log \sigma_k)^2, \quad
w_k = \begin{cases} 5.0 & \text{if } \log \sigma_k < 0 \text{ (compression)} \\ 1.0 & \text{otherwise (expansion)} \end{cases}
\label{eq:jac}
\end{equation}

This asymmetry reflects the observation that compression (cells crowding together) is more pathological than moderate expansion in tissue alignment~\cite{ref_jacreg}.

\subsection{Joint Optimization}

\paragraph{Spatial-Expression Alignment Loss (ours).}
Our core registration objective is an original \emph{spatial-expression alignment loss} that establishes correspondences between deformed source points $\hat{\mathbf{x}}^{(2)}$ and reference points $\mathbf{x}^{(1)}$ using a unified geometry--feature cost derived from the canonical field. For each deformed point, we find its $K$ nearest spatial neighbors in the reference and compute:
\begin{equation}
c_{ij} = \frac{\|\hat{\mathbf{x}}_j^{(2)} - \mathbf{x}_i^{(1)}\|^2}{s_{\text{sp}}} + \lambda_f \cdot \frac{1 - \langle \bar{\mathbf{e}}_j^{(2)}, \bar{\mathbf{e}}_i^{(1)} \rangle}{s_{\text{ft}}}
\end{equation}
where $\bar{\mathbf{e}} = \mathbf{e}/\|\mathbf{e}\|$ are L2-normalized ExprINR embeddings, $s_{\text{sp}}$ and $s_{\text{ft}}$ are running EMA scale factors, and $\lambda_f$ balances the two terms. Soft assignment weights are obtained as $p_{ij} = \text{softmax}(-c_{ij} / \tau)$, where $\tau$ is an adaptive temperature updated by EMA based on the mean weighted cost.

The embeddings entering the cost matrix are computed within a \texttt{torch.no\_grad()} context, keeping the canonical embeddings as a frozen prior during correspondence estimation and preventing degenerate feature collapse.

The matching loss is bidirectional: the forward loss pulls each deformed source point toward its weighted target centroid, while the reverse loss ensures each reference point is explained by at least one source point, preventing many-to-one collapse.

\paragraph{Reconstruction Loss.}
Following SUICA~\cite{ref_suica}, the expression decoder reconstructs HVG profiles from INR embeddings using a composite loss: masked MSE on nonzero entries, L1 on all entries, and a Dice loss on the zero/nonzero pattern:
\begin{equation}
\mathcal{L}_{\text{recon}} = \text{MSE}_{\text{masked}} + \text{L1} + 0.01 \cdot \mathcal{L}_{\text{dice}}
\end{equation}

\paragraph{Two-Phase Training.}
\textit{Phase~1} (300 epochs): Two independent ExprINRs ($f_{\theta_1}$ for the reference, $f_{\theta_2}$ for the source) share a single decoder and are jointly trained via $\mathcal{L}_{\text{recon}}$ on both slices, aligning their embedding spaces. Concurrently, the DeformationNet is independently pre-trained with PCA-based matching. The coarse-to-fine $\alpha$ window opens progressively over the first third of training. The best reconstruction checkpoint is restored before Phase~2.

\textit{Phase~2} (400 epochs): $f_{\theta_2}$ is discarded; all remaining components ($f_{\theta_1}$, $h_\phi$, $\phi_\psi$, and the matcher) are jointly optimized with the total loss:
\begin{equation}
\mathcal{L} = \mathcal{L}_{\text{match}} + \lambda_r \mathcal{L}_{\text{recon}} + \lambda_j \mathcal{L}_{\text{jac}}
\label{eq:total}
\end{equation}
where $\lambda_r = 0.1$ (reduced to prevent reconstruction from dominating alignment) and $\lambda_j$ controls Jacobian regularization. A full-coverage reverse matching step is performed each epoch to address global collapse missed by mini-batch training. Learning rate is scheduled via ReduceLROnPlateau.

\section{Experiments}

\subsection{Datasets and Baselines}

We evaluate nine ST datasets spanning grid-based and continuous-coordinate technologies: DLPFC~\cite{ref_slideseq} (3 samples, the first adjacent section pair per sample, ${\sim}$3.4--4.8k spots), STARMap~\cite{ref_starmap} (${\sim}$1k spots), MERFISH Brain~\cite{ref_zhang2021} (3 samples, ${\sim}$5.5--24.8k cells) and MERFISH Hypothalamus~\cite{ref_moffitt2018} (1 sample, ${\sim}$17--20k cells), and MouseEmbryo~\cite{ref_embryo} (${\sim}$17--20k cells). Alignment baselines are PASTE~\cite{ref_paste}, STalign~\cite{ref_stalign}, Spateo~\cite{ref_spateo}, plus Raw (unaligned) and ICP (rigid-only); integration-focused methods are evaluated separately in Table~\ref{tab:embedding} to avoid task mismatch. We report OT Accuracy~\cite{ref_paste}, NN Accuracy, and Chamfer distance~\cite{ref_chamfer,ref_benchmark} for alignment, and ARI/NMI (mclust) for embedding quality.

\subsection{Spatial Alignment Results}

\begin{table}[t]
\centering
\caption{D1--D3: DLPFC samples; MB2/MB7/MB11: MERFISH Brain sections; ST: STARMap; MH: MERFISH Hypothalamus; ME: MouseEmbryo. Spatial alignment results across nine datasets. Best in \textbf{bold}, second-best \underline{underlined}. ``--'' indicates method not available or no alignment.}
\label{tab:alignment}
\resizebox{\textwidth}{!}{%
{\fontsize{8}{9.5}\selectfont
\begin{tabular}{@{}l l ccc ccc c c c c@{}}
\toprule
& & \multicolumn{3}{c}{DLPFC} & \multicolumn{3}{c}{MERFISH Brain} & & & & \\
\cmidrule(lr){3-5}\cmidrule(lr){6-8}
Metric & Method & \rotatebox{45}{D1} & \rotatebox{45}{D2} & \rotatebox{45}{D3} & \rotatebox{45}{MB2} & \rotatebox{45}{MB7} & \rotatebox{45}{MB11} & \rotatebox{45}{ST} & \rotatebox{45}{MH} & \rotatebox{45}{ME} & \rotatebox{45}{Mean} \\
\midrule
\multirow{6}{*}{OT Acc\,$\uparrow$}
& Raw            & .745 & .887 & .673 & .013 & .050 & .209 & .860 & .051 & .116 & .400 \\
& PASTE          & .859 & \underline{.904} & .806 & .267 & .726 & \underline{.349} & \textbf{.882} & .674 & .031 & .611 \\
& STalign        & \textbf{.866} & \textbf{.905} & .808 & .022 & .046 & .143 & \underline{.861} & .051 & .112 & .424 \\
& Spateo         & \underline{.864} & .884 & .818 & \textbf{.343} & \textbf{.750} & .298 & .860 & .681 & \underline{.723} & \underline{.691} \\
& ICP            & .829 & .888 & \underline{.829} & .251 & .709 & .347 & .852 & \underline{.689} & .710 & .678 \\
& \textbf{Ours}  & .863 & .899 & \textbf{.839} & \underline{.318} & \underline{.746} & \textbf{.357} & .857 & \textbf{.690} & \textbf{.746} & \textbf{.702} \\
\midrule
\multirow{6}{*}{NN Acc\,$\uparrow$}
& Raw            & .765 & .913 & .729 & .130 & .064 & .276 & \underline{.890} & .167 & .172 & .456 \\
& PASTE          & \underline{.859} & .904 & .804 & .244 & .731 & \underline{.360} & .883 & .680 & .032 & .611 \\
& STalign        & \textbf{.875} & .894 & .788 & .136 & .065 & .181 & .872 & .167 & .166 & .460 \\
& Spateo         & .845 & \textbf{.925} & .823 & \textbf{.378} & \textbf{.784} & .215 & .889 & \textbf{.745} & \underline{.759} & \underline{.707} \\
& ICP            & .851 & \underline{.925} & \underline{.831} & .228 & .723 & .280 & .889 & \underline{.701} & .733 & .685 \\
& \textbf{Ours}  & .853 & .920 & \textbf{.849} & \underline{.322} & \underline{.780} & \textbf{.365} & \textbf{.899} & .693 & \textbf{.788} & \textbf{.719} \\
\midrule
\multirow{6}{*}{Chamfer\,$\downarrow$}
& Raw            & \underline{0.1} & \textbf{0.1} & 0.5 & -- & -- & -- & 136.5 & -- & -- & -- \\
& PASTE          & 0.3 & 0.3 & \textbf{0.2} & 173.1 & \textbf{5.8} & \underline{51.7} & \textbf{15.3} & \textbf{4.4} & 235.5 & 54.1 \\
& STalign        & 0.6 & 0.6 & 0.7 & -- & -- & -- & 140.6 & -- & -- & -- \\
& Spateo         & 0.6 & 0.7 & 0.7 & 166.2 & 16.8 & 103.3 & 142.7 & 12.2 & \underline{12.0} & 50.6 \\
& ICP            & \textbf{0.0} & \underline{0.1} & \underline{0.4} & \underline{77.9} & 18.7 & 107.0 & 135.7 & \underline{11.0} & 15.3 & \underline{40.7} \\
& \textbf{Ours}  & 0.7 & 0.6 & 0.6 & \textbf{12.6} & \underline{13.9} & \textbf{17.1} & \underline{135.0} & \underline{11.0} & \textbf{11.9} & \textbf{22.6} \\
\bottomrule
\end{tabular}
}}
\end{table}

Table~\ref{tab:alignment} summarizes alignment across nine datasets. INST-Align achieves the best mean OT Accuracy (0.702), NN Accuracy (0.719), and Chamfer distance (22.6), compared with Spateo (0.691/0.707/50.6). On near-aligned DLPFC (D1--D3), all methods are competitive after rigid initialization; our slightly higher Chamfer (0.6--0.7 vs.\ PASTE/ICP) reflects limited room for non-rigid improvement on already well-aligned grid data.

On large-deformation datasets the gap widens: Chamfer drops from 173.1 to 12.6 on MB2 (vs.\ PASTE), from 51.7 to 17.1 on MB11, and from 235.5 to 11.9 on MouseEmbryo, while OT/NN scores remain comparable. STalign produces excessively large deformations on MERFISH Brain and MouseEmbryo, making its metrics non-comparable (marked ``--''). Overall, the advantage of INST-Align grows with dataset size and deformation severity, suggesting that the canonical field provides stronger regularization as geometric ambiguity increases.

\subsection{Downstream Analysis}

\begin{table}[t]
\centering
\caption{Embedding quality (ARI / NMI) via mclust clustering. Each is from the target slice. Baselines include STAligner~\cite{ref_staligner}, GraphST~\cite{ref_graphst}, SPIRAL~\cite{ref_guo2023}, and Seurat~\cite{ref_seurat}.}
\label{tab:embedding}
\resizebox{\textwidth}{!}{
\setlength{\tabcolsep}{2.5pt}
{\fontsize{8}{9.5}\selectfont
\begin{tabular}{@{}l cc cc cc cc cc@{}}
\toprule
& \multicolumn{2}{c}{DLPFC Samp.~1} & \multicolumn{2}{c}{DLPFC Samp.~2} & \multicolumn{2}{c}{DLPFC Samp.~3} & \multicolumn{2}{c}{MouseEmbryo} & \multicolumn{2}{c}{Mean} \\
\cmidrule(lr){2-3}\cmidrule(lr){4-5}\cmidrule(lr){6-7}\cmidrule(lr){8-9}\cmidrule(lr){10-11}
Method & ARI & NMI & ARI & NMI & ARI & NMI & ARI & NMI & ARI & NMI \\
\midrule
PCA        & .298 & .385 & .205 & .364 & .208 & .340 & .128 & .215 & .210 & .326 \\
Seurat     & .237 & .372 & .285 & .287 & .253 & .397 & .078 & .220 & .213 & .319 \\
STAligner  & .273 & .474 & \underline{.463} & \underline{.556} & .293 & \underline{.491} & \textbf{.356} & \textbf{.552} & .346 & \underline{.518} \\
GraphST    & \underline{.431} & \underline{.641} & \textbf{.490} & \textbf{.589} & \textbf{.618} & \textbf{.724} & .266 & \underline{.516} & \textbf{.451} & \textbf{.618} \\
SPIRAL     & \textbf{.558} & \textbf{.652} & .359 & .484 & \underline{.356} & .478 & .192 & .355 & \underline{.366} & .492 \\
INST-Align (Ours) & .397 & .490 & \underline{.463} & .446 & .246 & .393 & \underline{.341} & .512 & .362 & .460 \\
\bottomrule
\end{tabular}
}}
\end{table}

Table~\ref{tab:embedding} reports embeddings from the canonical expression field. Because INST-Align is optimized for alignment, embedding quality is a byproduct rather than the primary target, yet results remain competitive (e.g., ARI 0.463 on DLPFC Sample~2, matching STAligner; ARI/NMI 0.341/0.512 on MouseEmbryo). GraphST achieves the highest mean ARI/NMI with dedicated spatial graph networks. That a coordinate-conditioned INR designed for alignment produces biologically meaningful embeddings without graph construction supports the view that a well-regularized canonical field captures tissue structure as an emergent property.
\begin{figure}[t]
\centering
\includegraphics[width=0.62\textwidth,trim=0 120 0 0,clip]{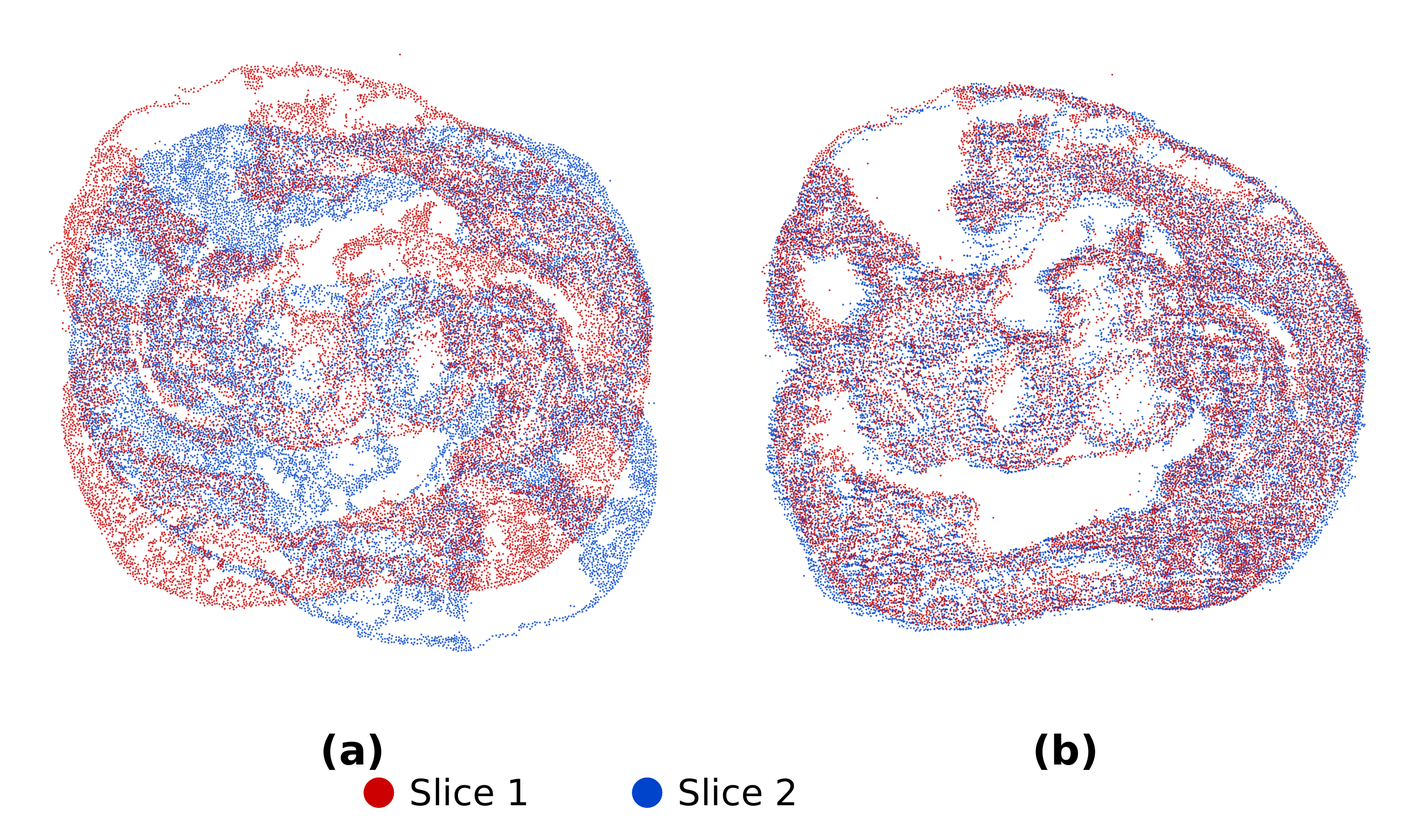}\hfill
\includegraphics[width=0.36\textwidth,trim=0 100 0 0,clip]{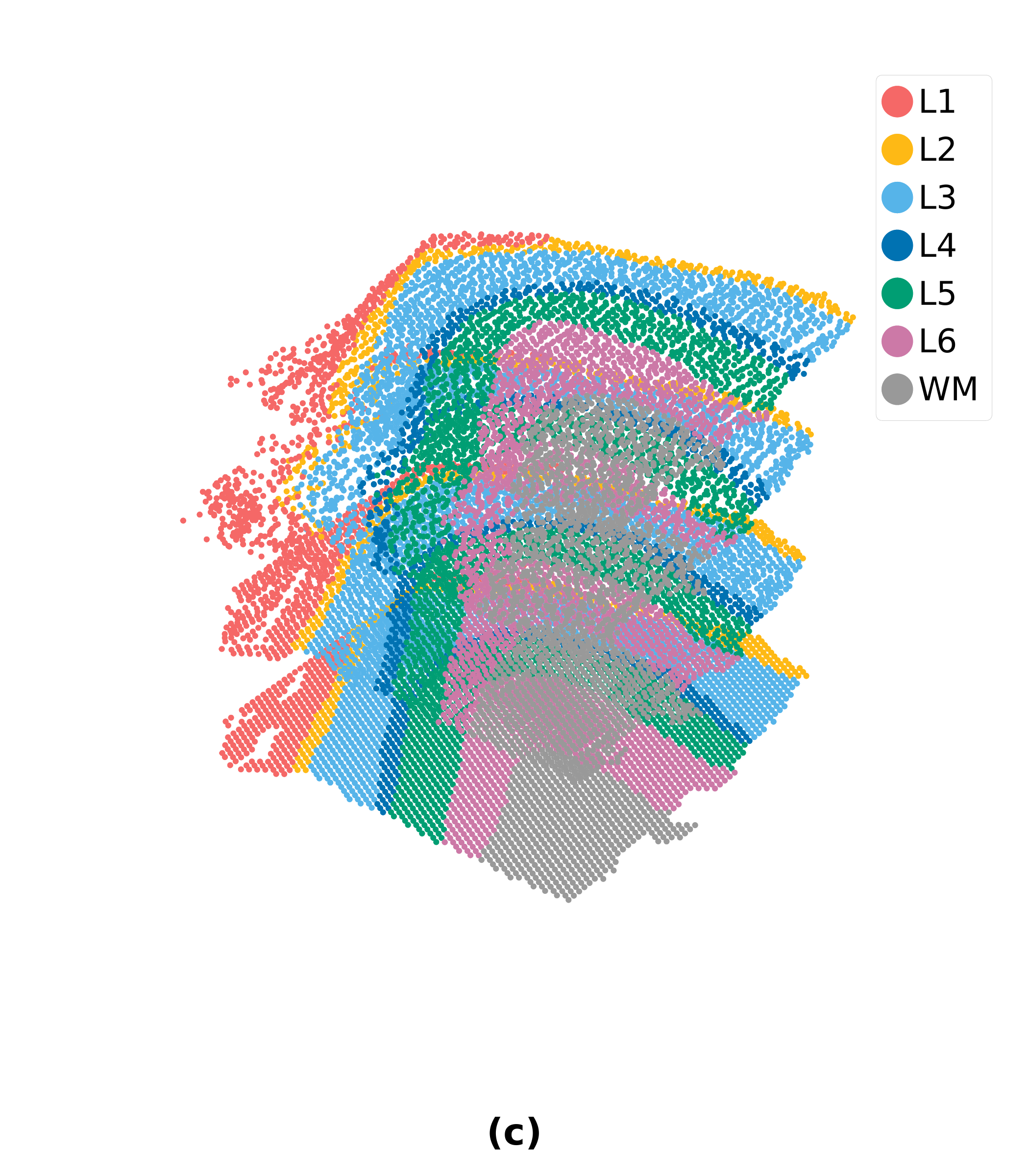}\\[2pt]
\makebox[0.62\textwidth]{\small\textbf{(a)} \textcolor{red}{$\bullet$} Slice~1 \quad \textcolor{blue}{$\bullet$} Slice~2 \quad\quad \textbf{(b)}}\hfill
\makebox[0.36\textwidth]{\small\textbf{(c)}}
\caption{Qualitative results. (a)~MouseEmbryo after rigid ICP pre-alignment; (b)~INST-Align non-rigid alignment on the same pair; (c)~3D reconstruction of DLPFC Sample~3 from consecutively aligned slices, colored by cortical layer.}
\label{fig:qualitative}
\end{figure}

Fig.~\ref{fig:qualitative} shows that INST-Align resolves large non-rigid mismatches on MouseEmbryo~(a$\to$b) and produces coherent layer structure in 3D reconstruction~(c).

\subsection{Ablation Studies}

\begin{table}[t]
\centering
\caption{Ablation study on DLPFC Sample~1. Each row removes one component from the full model.}
\label{tab:ablation}
\setlength{\tabcolsep}{5pt}
{\fontsize{8}{9.5}\selectfont
\begin{tabular}{@{}l cccc@{}}
\toprule
& \multicolumn{2}{c}{Alignment} & \multicolumn{2}{c}{Embedding} \\
\cmidrule(lr){2-3}\cmidrule(lr){4-5}
Variant & OT Acc\,$\uparrow$ & Chamfer\,$\downarrow$ & ARI\,$\uparrow$ & NMI\,$\uparrow$ \\
\midrule
Full (Ours)              & .810          & \textbf{0.36} & \textbf{.334} & \textbf{.450} \\
w/o Phase~1 (pretrain)   & \textbf{.813} & 0.47          & .045          & .079          \\
w/o Phase~2 (joint opt.) & .807          & 0.56          & .295          & .434          \\
w/o Jacobian reg.        & .805          & 0.49          & .287          & .400          \\
\bottomrule
\end{tabular}
}
\end{table}

Table~\ref{tab:ablation} shows that pretraining is critical for embedding quality (ARI drops from 0.334 to 0.045 without it), joint optimization improves geometric alignment (Chamfer 0.36 vs.\ 0.56), and Jacobian regularization benefits both.

\section{Conclusion}

We presented INST-Align, which couples a deformation network with a shared Canonical Expression Field for pairwise ST alignment and reconstruction. The shared field provides biological regularization for ambiguous correspondences under sparsity, batch effects, and large deformations. Across nine datasets, INST-Align achieves state-of-the-art mean OT Accuracy, NN Accuracy, and Chamfer distance, with consistent performance across both near-aligned grid data and large-deformation sections, yielding Chamfer reductions of up to 94.9\% relative to PASTE. Ablations confirm contributions from Phase~1 pretraining, Phase~2 joint optimization, and Jacobian regularization. The current framework operates in a pairwise setting and lacks a global multi-slice consistency objective. In addition, biological validation is limited to clustering-based metrics (ARI/NMI), and the learned embeddings remain below specialized embedding methods. Future work will extend the model toward globally consistent multi-slice 3D alignment with enhanced biological evaluation.

\end{document}